\documentclass[a4paper,11pt]{article}

\usepackage{fourier}
\usepackage{color}
 \usepackage{graphicx}
\usepackage{url}
\usepackage[affil-it]{authblk}
\usepackage{amsmath}
\usepackage{wrapfig}

\usepackage[T1]{fontenc}
\usepackage{times}
\usepackage{cite}
\usepackage{amsmath,amssymb,amsfonts}
\usepackage{algorithmic}
\usepackage{graphicx}
\usepackage{textcomp}
\usepackage{xcolor}
\usepackage{subfig}

\begin{document}

\title{ OxML Challenge 2023: Carcinoma classification using data augmentation}

\author[1]{Kislay Raj}
\author[1]{Teerath  Kumar }
\author[2]{Alessandra Mileo }
\author[3]{Malika Bendechache }

\affil[1]{CRT-AI Centre,
School of Computing,
Dublin City University, Ireland}

\affil[2]{INSIGHT Research Centre,
School of Computing,
Dublin City University, Ireland}

\affil[3]{Lero Research Centre,
School of Computer Science,
University of Galway, Ireland}

\date{}
\maketitle
\thispagestyle{empty}

\begin{abstract}
Carcinoma is the prevailing type of cancer and can manifest in various body parts. It is widespread and can potentially develop in numerous locations within the body. In the medical domain, data for carcinoma cancer is often limited or unavailable due to privacy concerns. Moreover, when available, it is highly imbalanced, with a scarcity of positive class samples and an abundance of negative ones. The OXML 2023 challenge provides a small and imbalanced dataset, presenting significant challenges for carcinoma classification. To tackle these issues, participants in the challenge have employed various approaches, relying on pre-trained models, preprocessing techniques, and few-shot learning. Our work proposes a novel technique that combines padding augmentation and ensembling to address the carcinoma classification challenge.  In our proposed method, we utilize ensembles of five neural networks and implement padding as a data augmentation technique,   taking into account varying image sizes to enhance the classifier's performance. Using our approach, we made place into top three and declared as winner.
\end{abstract}

\section{Introduction}
Carcinoma is a widespread and complex cancer originating from epithelial cells lining various organs and tissues, including the skin, lungs, breast, prostate, liver, and kidneys. Detailed cases and deaths in 2020 are documented in~\cite{gupta2022deep}. Its diverse presentations and early detection challenges make carcinoma a critical area of study for medical researchers~\cite{ranjbarzadeh2023me}, as early detection can save many lives. 

Deep learning (DL) has achieved success in various domains such as image~\cite{aleem2022random,singh2023efficient,singh2024efficient,raj2023neuro,kumarforged,kumar2021binary,vavekanand2024cardiacnet,kumar2024keeporiginalaugment}, audio~\cite{fu2010survey,park2020search,kumar2020intra,kumar2023audrandaug}, and text~\cite{torfi2020natural}. However, medical images are scarce, and available data often needs balancing. DL models require substantial, balanced datasets for good generalization. DL techniques have shown promise in carcinoma detection, enhancing accuracy, early diagnosis, and personalized treatment. Convolutional Neural Networks (CNNs) are particularly effective in extracting detailed patterns from medical images, supporting reliable cancer detection~\cite{boveiri2020medical}.

Recently, the OXML competition for carcinoma classification~\cite{oxml-carinoma-classification} attracted several teams. The competition aims to categorize hematoxylin and eosin-stained histopathological slices into two groups: those with carcinoma cells and those without. If carcinoma is present, it is further classified as malignant or not, resulting in three distinct classes. The competition presents several challenges:


\begin{itemize}
    \item Training data is very small; only 62 training images were provided.
    \item Size of all the images is different. There are risks associated with cropping images, as it may lead to the omission of target cells. Similarly, resizing images can alter their features and potentially reduce their readability.
    \item Dataset is heavily imbalanced, so model can be biased toward the majority class. 
\end{itemize}


To address these challenges, we present a novel approach that combines the power of deep learning with advanced image preprocessing techniques, such as data augmentation~\cite{kumar2023advanced}, which increases the generalization capability of the model, deals with class imbalance issues  and increases the diversity of data by increasing the size of the dataset,  and normalization, to enhance the performance of carcinoma cancer classification~\cite{wang2020comparison}.

The rest of the paper is organized as follows: section~\ref{Methodology} explains our used methodology, section~\ref{experiments} discusses the results and section~\ref{conclusion} concludes the work. 
\begin{figure}[htbp]
    \centering
    \includegraphics[height=5cm, width=0.50\textwidth]{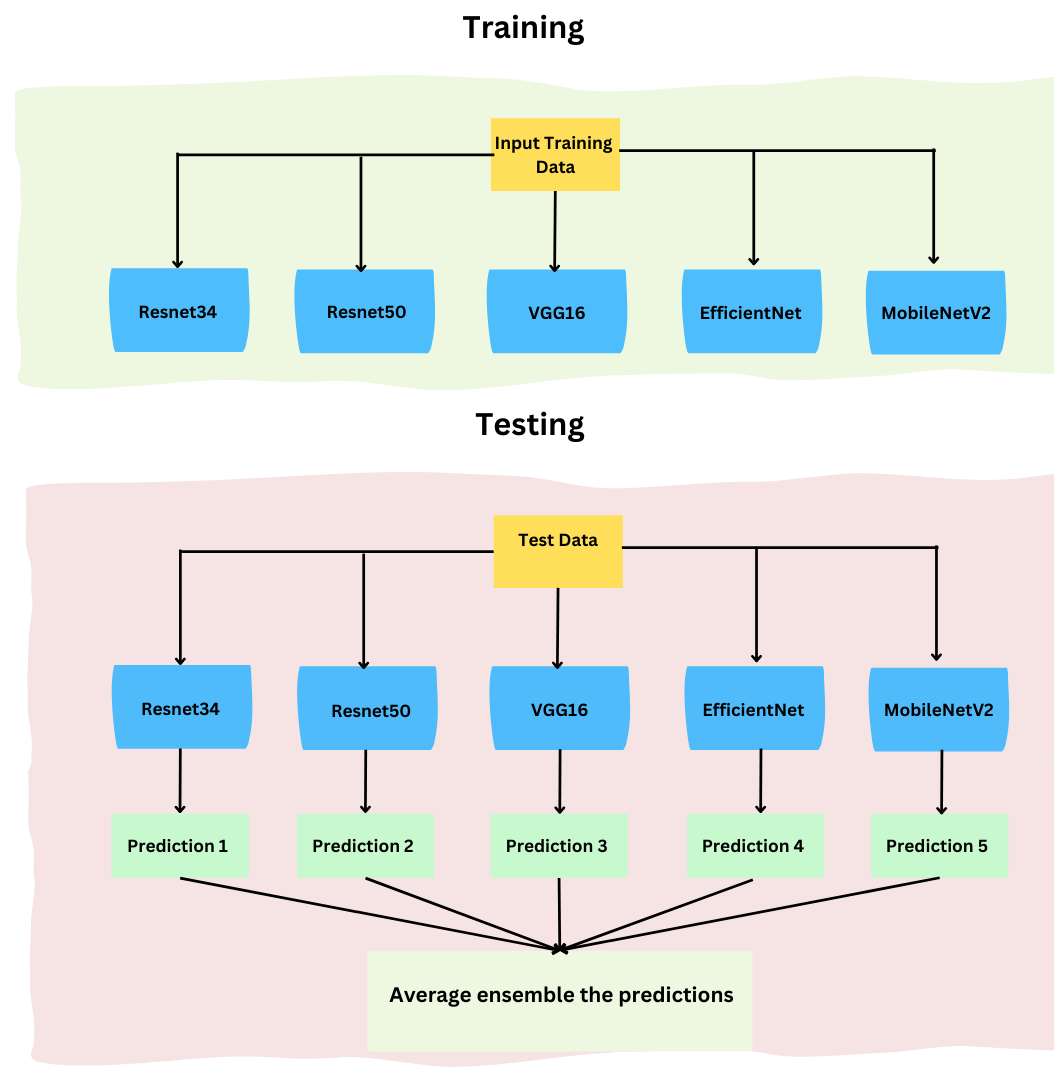}
    \caption{Training and testing using ensemble technique}
    \label{fig:ensembling_flow}
\end{figure}

\section{Methodology}\label{Methodology}
In our methodology, data augmentation and ensembling technique so we divide this section into two subsections: 
\subsection{Data Augmentation}


In the OxML competition, images varied in size, making it impractical to resize them uniformly without losing important features. Instead, we preserved the original sizes by padding them to the dimensions of the largest image ensuring feature retention and target cell visibility. We also faced a significant class imbalance in the dataset of 62 training images: 36 samples for class -1 , 14 for class 0 , and 12 for class 1. To address this, we used jitter data augmentation to increase the samples for the minority classes (0 and 1) to match the majority class (-1).

\subsection{Ensemble learning}


Ensemble learning enhances prediction performance by combining outputs from multiple models, improving generalization and accuracy, especially in fields like medicine where data is often scarce~\cite{aleem2023ensemble, hameed2020breast, singha2022ensemble}.

In our study, we use an ensemble of five diverse models: ResNet34, ResNet50, VGG16, EfficientNet, and MobileNetV. Each model is trained on the available training data. During testing, we feed the test data to each model and average their predictions. This process is shown in figure~\ref{fig:ensembling_flow}.
\section{Experiments}\label{experiments}
\subsection{Dataset}
The dataset provided for the OXML Carcinoma Classification task consisted of a total of 186 images. Among these, 62 images were labeled and used for training purposes, while another set of 124 images were allocated for validation (public score). However, these 124 validation images were not provided to the participants, as they were reserved for testing purposes (private score)~\cite{oxml-carinoma-classification}.
\subsection{Experimental setup}

In our approach, we employ five distinct pre-trained models, and for each model, we conduct training for 100 epochs. To ensure maximum flexibility, we enable the training of all layers within each model. We utilize stochastic gradient descent (SGD) as our optimizer for optimisation, with a learning rate of 0.001 and a momentum value of 0.9. We picked the best model based on optimal loss and saved the model. However, F1-score, a metric that provides a balanced trade-off between sensitivity (recall) and specificity, is used in the competition.

\begin{table}[htbp]
    \centering
    \begin{tabular}{|c|c|c|}
        \hline
         \textbf{Team Name} & \textbf{Public Score} &  \textbf{Private Score} \\  \hline 
         DCU CRT-AI & 0.79023 & 0.7258\\  \hline 
         Jessy &  0.79032 & 0.74193\\  \hline 
         Fatih Aksu & 0.77419 & 0.74193 \\  \hline 
         Minerva's Data Lab & 0.82258
 &  0.72580  \\  \hline 
    \end{tabular}
    \caption{OxML Health Track winners score}
    \label{tab:f1_scores}
\end{table}
\subsection{Results}
The competition attracted participation from a total of 39 teams. Among these teams, the top three performers were declared as the winners. Each team employed its unique approach to tackle the problem. However, for the purpose of this study, we focus on reporting the public and private scores achieved by only the winning teams. These scores are key indicators of the success and effectiveness of their respective methodologies in the competition. Top three teams' score are report in table~\ref{tab:f1_scores}.

\section{Conclusion}\label{conclusion}



In this study, we tackled carcinoma classification using a small, imbalanced dataset by combining padding data augmentation with ensemble learning. We employed five neural network models, ensuring each layer was trainable, and maintained consistent image sizes through padding augmentation. Our method showed promising results in the OxML Challenge 2023: Carcinoma Classification ML x Health Track, which featured 39 teams, with the top three determined by public and private scores. Our approach, along with others, improved carcinoma classification performance, demonstrating the value of the competition as a platform for advancing this field. As research progresses, continued collaborations will further enhance cancer detection and medical image analysis.

\section*{Acknowledgment}
We want to acknowledge the organizers of the OxML Challenge 2023 for providing the dataset and creating a competitive environment that fosters innovation and collaboration in medical image analysis. This work was conducted with the financial support of the Science Foundation Ireland Centre for Research and Training in Artificial Intelligence under Grant No. 18/CRT/6223.
\par Finally we thank to Dr. Surendar Nangani, Dr. NajeebUllah and Dr. Aneel Kumar from Pakistan for their medical domain insights and Vincent Moens, Research Scientist at Meta AI, London England for hosting this competition and providing dataset.

\bibliographystyle{apalike}

\bibliography{imvip}

\end{document}